\documentclass[review]{elsarticle}

\usepackage{lineno,hyperref}
\usepackage{amssymb,amsmath,amsthm}
\usepackage{cleveref}
\usepackage{color}
\usepackage{subfigure}
\usepackage{bm}
\usepackage{framed}
\usepackage{booktabs}
\usepackage{color}

\modulolinenumbers[5]
\definecolor{shadecolor}{rgb}{0.92,0.92,0.92}
\allowdisplaybreaks[2]

\journal{}

\usepackage{numcompress}

\begin{document}

\begin{frontmatter}

\title{UniG-Encoder: A Universal Feature Encoder for Graph and Hypergraph Node Classification}

%% or include affiliations in footnotes:
\author[mymainaddress]{Minhao Zou}

\author[mymainaddress]{Zhongxue Gan\corref{mycorrespondingauthor}}
\ead{ganzhongxue@fudan.edu.cn}

\author[mymainaddress]{Yutong Wang}

\author[mymainaddress]{Junheng Zhang}

\author[mymainaddress]{Dongyan Sui}

\author[mymainaddress]{Chun Guan\corref{mycorrespondingauthor}}
\ead{chunguan@fudan.edu.cn}

\author[mymainaddress,iics]{Siyang Leng\corref{mycorrespondingauthor}}
\cortext[mycorrespondingauthor]{To whom correspondence should be addressed.}
\ead{syleng@fudan.edu.cn}

\address[mymainaddress]{Institute of AI and Robotics, Academy for Engineering and Technology, Fudan University, Shanghai 200433, China}
\address[iics]{Research Institute of Intelligent Complex Systems, Fudan University, Shanghai 200433, China}

\begin{abstract}
Graph and hypergraph representation learning has attracted increasing attention from various research fields. Despite the decent performance and fruitful applications of Graph Neural Networks (GNNs), Hypergraph Neural Networks (HGNNs), and their well-designed variants, on some commonly used benchmark graphs and hypergraphs, they are outperformed by even a simple Multi-Layer Perceptron. This observation motivates a reexamination of the design paradigm of the current GNNs and HGNNs and poses challenges of extracting graph features effectively. In this work, a universal feature encoder for both graph and hypergraph representation learning is designed, called UniG-Encoder. The architecture starts with a forward transformation of the topological relationships of connected nodes into edge or hyperedge features via a normalized projection matrix. The resulting edge/hyperedge features, together with the original node features, are fed into a neural network. The encoded node embeddings are then derived from the reversed transformation, described by the transpose of the projection matrix, of the network's output, which can be further used for tasks such as node classification. The proposed architecture, in contrast to the traditional spectral-based and/or message passing approaches, simultaneously and comprehensively exploits the node features and graph/hypergraph topologies in an efficient and unified manner, covering both heterophilic and homophilic graphs. The designed projection matrix, encoding the graph features, is intuitive and interpretable. Extensive experiments are conducted and demonstrate the superior performance of the proposed framework on twelve representative hypergraph datasets and six real-world graph datasets, compared to the state-of-the-art methods. Our implementation is available online at \url{https://github.com/MinhZou/UniG-Encoder}.
\end{abstract}


\end{frontmatter}

%\linenumbers

\section{Introduction}
% and learn node-level and graph-level representations
Graph and hypergraph representation learning is a rapidly growing field of research that focuses on learning meaningful representations from nodes and edges/hyperedges features in graph/hypergraph-structured data. This field has seen significant progress in recent years due to the development of advanced techniques such as Graph Neural Networks (GNNs) and Hypergraph Neural Networks (HGNNs), which are capable of modeling complex interactions in real-world scenarios. Particularly, HGNNs are designed to extend GNNs to capture higher-order relationships among more than two nodes, which are ubiquitous in social networks~\cite{patania2017shape,qiu2023closed}, ecological networks~\cite{bairey2016high}, biological networks~\cite{petri2014homological}, etc. A fundamental task in graph/hypergraph representation learning is node classification that categorizing nodes based on their features and graph/hypergraph topologies.

% which or applying data augmentation techniques such as feature shuffle, node dropping, and node interpolation \cite{xue2021node}
Most of the existing literatures stick to learning node embeddings from neighbors using powerful neural operators, such as convolution~\cite{kipf2016semi, ma2019graph, yadati2019hypergcn}, attention~\cite{velickovic2017graph, georgiev2022heat,zou2023similarity}, spectrum~\cite{zien1999multilevel, sun2008hypergraph}, and diffusion~\cite{wang2022equivariant}. These approaches have resulted in the popular spectral-based and message passing architectures~\cite{huang2021unignn, feng2019hypergraph, gao2022hgnn, chien2021you}. Despite their wide applications, these approaches have limitations, as a simple Multi-Layer Perceptron (MLP) can even outperform well-designed GNNs, HGNNs, and their variants on some commonly used benchmark graphs and hypergraphs, see the results in Table \ref{table: acc-1} for \texttt{Zoo}, \texttt{House}, \texttt{Senate}, \texttt{Cornell}, \texttt{Texas}, and \texttt{Wisconsin} datasets. The major drawback of the spectral-based architecture is its heavy reliance on the homophily assumption, which requires that nodes with similar features and/or labels tend to be linked. The message passing architecture conducts aggregation on the raw node embeddings or considers only the node-to-edge then edge-to-node mapping procedure, which can lead to suboptimal performance in some cases. To address these issues, a new approach, called UniG-Encoder, is proposed which simultaneously and comprehensively exploits the node features and graph/hypergraph topologies.

Drawing inspiration from Hypergraph Line Expansion~\cite{yang2022semi}, which treats nodes and edges equally and converts hyperedges into ``line nodes'', our architecture leverages these approaches by treating edges/hyperedges as additional nodes and extracting their features from the topological relationships of the connected nodes. Edges/hyperedges that connect two or more nodes are transformed into additional feature vectors, enabling tuning the weights between node features and graph structure based on the homophilic extent thus alleviating the curse of heterophily. This is efficiently accomplished by using a normalized projection matrix, linearly combining the features of connected nodes and resulting the edge/hyperedge features. These generated features, together with the original node features, are fed into a neural network, e.g., MLP, Transformer~\cite{vaswani2017attention}, etc., and its output is processed via a reversed transformation, aggregating neighborhood features by the transpose of the projection matrix, to obtain the encoded node embeddings, which can be further used for tasks such as node classification. The proposed framework is demonstrated by extensive experiments to outperform the state-of-the-art methods on eighteen benchmark datasets with diverse properties. We summarize the main contributions of our work as:
\begin{itemize}
\item A universal framework UniG-Encoder is proposed towards representation learning for both graphs and hypergraphs, covering also both heterophilic and homophilic circumstances by leveraging simultaneously the information of node features and topology.
\item The architecture is realized via an intuitive and interpretable normalized projection matrix, enabling tuning the weights between node features and graph structure based on the homophilic extent, which can be easily acquired from \textit{a priori} knowledge of datasets.
\item The designed architecture involves minor computation consumption but achieves superior performance over the state-of-the-art methods on representative datasets, supported by extensive analysis and experiments.
\end{itemize}

\section{Related Works}

\textbf{Graph and Hypergraph Neural Networks.} GNNs and HGNNs learn informative graph/hypergraph embeddings by leveraging the node features and structure. Various variants of GNNs and HGNNs have been developed, and we review the most recent advances here.

Spectral-based approaches interpret graph convolution from the perspective of graph signal processing, with the aim of removing noise from graph signals or smoothing information among connected nodes. GCN~\cite{kipf2016semi} applies convolutional operation in the spectral domain to input features, generating node embeddings for node classification and other downstream tasks. Building upon GCN, GCNII~\cite{chen2020simple} employs initial residual and identity mapping to effectively alleviate the over-smoothing problem. The spectral-based approach has also been extended to deal with hypergraphs, such as HyperGCN~\cite{yadati2019hypergcn}.

Spatial-based approaches aggregate messages from neighboring nodes via message passing layers, known as Message Passing Neural Networks (MPNNs)~\cite{gilmer2017neural}. GraphSAGE~\cite{hamilton2017inductive} generates node embeddings by aggregating information from a fixed number of neighbors. In contrast, GAT~\cite{velickovic2017graph} uses an attention mechanism to weigh the contributions of neighboring nodes and aggregates information from these neighbors based on learned weights. Many GNNs have also been developed for heterophilic problem, such as H2GCN~\cite{zhu2020beyond}, GGCN~\cite{yan2022two}, and GloGNN~\cite{li2022finding}. In hypergraph representation learning, many works use a two-stage message passing process, such as HGNN~\cite{feng2019hypergraph}, AllSet (AllDeepSets and AllSetTransformer)~\cite{chien2021you}, and UniGCNII~\cite{huang2021unignn}. In most cases when the number of edges is significantly larger than the number of nodes, these two-stage methods suffer from computational burden due to enormous intermediate edge embeddings.

\textbf{Hypergraph Expansion.} In the realm of hypergraph analysis, a technique known as hypergraph expansion is often used to transform hypergraphs into graphs. One prominent algorithm for hypergraph expansion is the clique expansion~\cite{sun2008hypergraph}, which generates a graph from hypergraph by substituting each hyperedge with a clique in the resulting graph. Another approach, known as the star expansion algorithm~\cite{zien1999multilevel}, creates a graph by introducing a new vertex for every hyperedge, which is connected to each vertex in the hyperedge. Line expansion~\cite{yang2022semi} simplifies the hypergraph by treating nodes and hyperedges as equivalent, representing each vertex-hyperedge incident pair as a ``line node''. The expansion methods also bring in additional computational burden due to an extra number of edges expanded from hyperedges.

\section{Preliminaries}
% We introduce the definitions and notations related to GNNs/HGNNs in this section.
In this section, essential concepts, definitions, and notations pertaining to GNNs and HGNNs are presented.

A universal representation learning framework for both graphs and hypergraphs is proposed in this work, so we adopt a unified representation for them here. Both edges and hyperedges are defined as subsets of nodes, while an edge is a subset with two elements and a subset of hyperedge contains more than two nodes. Therefore, let $\mathcal{G}=(\mathcal{V},\mathcal{E})$ denote a graph or hypergraph, where $\mathcal{V}$ is the set of all nodes, and $\mathcal{E}$ is the set of edges or hyperedges defined above.

For an arbitrary set $\mathcal{S}$, the cardinality of it is denoted by $|\mathcal{S}|$. A graph or hypergraph $\mathcal{G}$ can be characterized by a $|\mathcal{V}|\times|\mathcal{E}|$ incidence matrix $\mathbf{B}$, where $\mathbf{B}_{ij}={\begin{cases}{1,}&{{\mathrm{if~}}v_i\in e_j}\\{0,}&{{\mathrm{otherwise}}}\end{cases}}$ with node $v_i\in\mathcal{V}$ and edge/hyperedge $e_j\in\mathcal{E}$. For $v_i\in\mathcal{V}$ and $e_j\in\mathcal{E}$, their degrees are defined as $d(v_i)=\sum\limits_{j}\textbf{B}_{ij}$ and $\delta(e_j)=\sum\limits_{i}\mathbf{B}_{ij}$, respectively. $\mathbf{D}_\mathcal{V}\in\mathbb{R}^{|\mathcal{V}|\times|\mathcal{V}|}$ and $\mathbf{D}_\mathcal{E}\in\mathbb{R}^{|\mathcal{E}|\times|\mathcal{E}|}$ denote the diagonal matrices of node degrees and edge degrees, respectively. The raw node features is described by matrix $\mathbf{X}\in\mathbb{R}^{|\mathcal{V}|\times{C_0}}$, where the $i$-th row vector $\mathbf{x}_i$ denotes the {\it ego-feature} of node $v_i$ and ${C_0}$ is the dimension of features.

\section{Methods}

\begin{figure*}[htbp]
  \centering
  \includegraphics[width=1.0\textwidth]{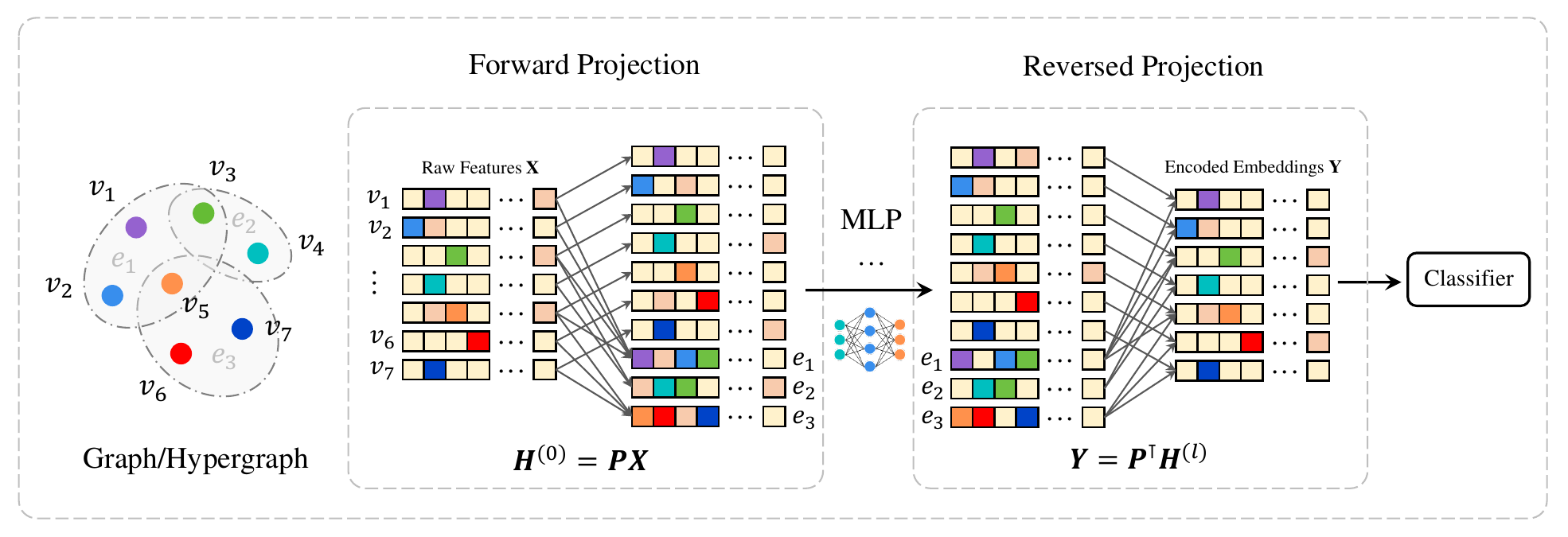}
  \caption{\textbf{The architecture of UniG-Encoder.} The architecture starts with a forward transformation of the topological relationships of connected nodes into edge or hyperedge features via a normalized projection matrix. The resulting edge/hyperedge features, together with the original node features, are fed into a neural network. The encoded node embeddings are then derived from the reversed transformation, described by the transpose of the projection matrix, of the network's output, which can be further used for tasks such as node classification. Notably, the MLP can be substituted by advanced neural networks such as the Transformer.}
  \label{fig: architecture}
\end{figure*}

We illustrate here the general architecture of UniG-Encoder for both graph and hypergraph representation learning. The key component lies in a normalized projection matrix that first forwardly converts the topological relationships of connected nodes into edge or hyperedge features. The resulting edge/hyperedge features, together with the original node features, are fed into a neural network. In this work, we use a simple MLP to process these features. The encoded node embeddings are then derived from the reversed transformation, described by the transpose of the projection matrix, of the MLP's output, which are subsequently used for node classification task. The architecture of UniG-Encoder is summarized in Figure \ref{fig: architecture}, with the detailed components described in the following subsections.

\subsection{Forward Projection}

\textbf{Projected Set.} Let $\mathcal{V}_P$ denote the set induced by the projection matrix from the graph/hypergraph $\mathcal{G}=(\mathcal{V},\mathcal{E})$, which is an ordered set consisting of two parts. The first part is a permutation of the node set $\mathcal{V}$ and the second part denotes the transformed edge/hyperedge set, while the features of the projected set are obtained from the raw node features with the projection matrix $\mathbf{P}$ acting on them.
% diagonal

\textbf{Projection Matrix.} We introduce a seminal version of the projection matrix here, whereas it can be redesigned to accommodate different homophilic extents of the graphs/hypergraphs. The projection matrix $\mathbf{P}$ is the row-wise concatenation of two matrices: the node part $\mathbf{P}_\mathcal{V}$ and the edge/hyperedge part $\mathbf{P}_\mathcal{E}$. The former is of size $\mathbb{R}^{|\mathcal{V}|\times|\mathcal{V}|}$, which is a column-wise permutation of the unit matrix $\mathbf{I}$. The latter $\mathbf{P}_\mathcal{E}$ is of size $\mathbb{R}^{|\mathcal{V}_\mathcal{E}|\times|\mathcal{V}|}$, whose element $\mathbf{P}_{\mathcal{E},ij}$ equals to $1$ if $v_j\in e_i$ and whose other elements are all zero. Therefore, $\mathbf{P}=\left[\begin{matrix}\mathbf{P}_\mathcal{V}\\\mathbf{P}_\mathcal{E}\end{matrix}\right]\in\mathbb{R}^{|\mathcal{V}_{P}|\times|\mathcal{V}|}$. The above forward projection procedure satisfies the following theorem:

\textbf{Theorem 1.} Assume there is no duplicate edge/hyperedge in a given graph/hypergraph $\mathcal{G}=(\mathcal{V},\mathcal{E})$. One can construct a map $\phi:\mathcal{V}\cup\mathcal{E}\to\mathcal{V}_P$, which is bijective under the construction according to the above corresponding relations as the projection matrix $\mathbf{P}$, and its inverse exists. This means that each node and edge/hyperedge in $\mathcal{G}$ is uniquely mapped to an element in $\mathcal{V}_P$, and vice versa.

% 满射的逆映射存在，需要证明该映射是可逆的

% # 双射的好处 减少冗余 减少计算

% \textbf{Theorem 4.} The mapping $\phi$ obtained using the projection matrix is a generalization of line expansion and therefore a generalization of clique expansion, star expansion, and simple graph convolution when $w=0$ (i.e., no message passing from hyperedge-similar neighbors). % 更加灵活 可以用到不同地方 维度也不一样 并且无需多余聚合操作 如利用GCN

\subsection{Feature Projection}

The projection matrix $\mathbf{P}$ in fact provides a new method for generating edge/hyperedge features by linearly combining the features of connected nodes, called feature projection. These resulting features, together with the original node features, are subsequently fed as input to an MLP to obtain new embeddings. Notably, the MLP can be replaced by other neural network architectures, such as RNN or Transformer, thereby enhancing the flexibility and adaptability of out framework in different application scenarios.

Concretely, for a graph/hypergraph $\mathcal{G}=(\mathcal{V},\mathcal{E})$, the projection matrix $\mathbf{P}$ is acted on the original node features $\mathbf{X}\in\mathbb{R}^{|\mathcal{V}|\times{C_0}}$, yielding the feature vectors of the projected set $\mathcal{V}_P$, i.e., $\mathbf{H}^{(0)}=\mathbf{P}\mathbf{X}\in\mathbb{R}^{|\mathcal{V}_{P}|\times C_{0}}$. In fact, $\mathbf{H}^{(0)}$ is the concatenation of the permutated node \textit{ego-features} and the generated edge/hyperedge features. Subsequently, $\mathbf{H}^{(0)}$ is fed into an $l$-layer neural network and the embedding for its $k$-th layer is denoted by $\mathbf{H}^{(k)}\in\mathbb{R}^{|\mathcal{V}_P|\times C_k} (k=1,\cdots,l)$.

\subsection{Reversed Projection}

The output of the neural network $\mathbf{H}^{(l)}$ is then reversely transformed by the transpose of the projection matrix, i.e., $\mathbf{P}^\top$. In fact, $\mathbf{P}^\top=\left[\mathbf{P}^\top_\mathcal{V},\mathbf{P}^\top_\mathcal{E}\right]\in\mathbb{R}^{|\mathcal{V}|\times|\mathcal{V}_{P}|}$. Thus the encoded node embeddings used for classification can be obtained by $\mathbf{Y}=\mathbf{P}^{\top}\mathbf{H}^{(l)}\in\mathbb{R}^{|\mathcal{V}|\times C}$, where $C=C_l$ denotes the number of final features. The resulting rows of $\mathbf{Y}$ are obtained by taking a weighted summation of the corresponding rows of $\mathbf{H}^{(l)}$, where the weights are given by the nonzero elements in each row of $\mathbf{P}^\top$. Here, the matrix $\mathbf{P}^{\top}_\mathcal{V}$ is used to extract representations from the \textit{ego-embeddings}, while $\mathbf{P}^{\top}_\mathcal{E}$ is used to extract representations from the edge/hyperedge embeddings. This operation represents an extension of the aggregation process from neighbors in message passing architecture or the spectral filter in spectral-based architecture, which simultaneously leverages both node embeddings and edge/hyperedge embeddings. The procedures satisfy the following theorem:

\textbf{Theorem 2.} Let $\sigma:\mathcal{V}\to\mathcal{V}$ be an arbitrary permutation, thus $\mathbf{P}_{\mathcal{V},ij}=\delta_{j,\sigma(i)}$, where $\delta_{ij}=\begin{cases}{1,}&{{\mathrm{if~}}i=j}\\{0,}&{{\mathrm{otherwise}}}\end{cases}$. It can be concluded that $(\mathbf{P}^\top\mathbf{P})_{ij}>0$ if and only if $i=j$ or $\exists e\in\mathcal{E}$ such that $v_i\in e$ and $v_j\in e$.

This theorem guarantees the exact correspondence during the projection process, that is, the encoded embeddings for node $v_i$ exactly contains the information from the raw features of node $v_i$ and the features of edges/hyperedges containing $v_i$.

\subsection{Normalization}

The defined projection matrix and its transpose are normalized in this framework. During the forward projection, the rows of $\mathbf{P}_\mathcal{E}\mathbf{X}$ are obtained by taking a weighted summation of the corresponding rows of $\mathbf{X}$, where the weights are given by the nonzero elements in each row of $\mathbf{P}_\mathcal{E}$. The intuition lies in that $\mathbf{P}_\mathcal{E}$ is used to fuse the features of connected nodes into the features of edges/hyperedges. Therefore, row normalization needs to be performed on $\mathbf{P}_\mathcal{E}$, i.e.,
\begin{equation}
\hat{\mathbf{P}}_{\mathcal{E},ij}=\frac{\mathbf{P}_{\mathcal{E},ij}}{\sum^{|\mathcal{V}|}_{k=1}\mathbf{P}_{\mathcal{E},ik}}.
\end{equation}
Row normalization is also used for the reversed transformation described by $\mathbf{P}^\top$, i.e.,
\begin{equation}
\hat{\mathbf{P}}^{\top}_{ij}=\frac{\mathbf{P}^{\top}_{ij}}{\sum^{|\mathcal{V}_P|}_{k=1}\mathbf{P}^{\top}_{ik}}.
\end{equation}
After normalization, the embeddings of nodes and the embeddings of edges/hyperedges they belong to are fused by weighted summation into the final encoded embeddings. We also try different normalization methods that trading-off the weights between node features and edge/hyperedge features in the experiments and compare their impacts.

\section{Experiments}

\subsection{Datasets}

Our framework is designed to accommodate both graphs and hypergraphs. To demonstrate its effectiveness, extensive experiments are conducted on various benchmark datasets. We briefly introduce the datasets used in this work, with their detailed information listed in Appendix.

\textbf{Graph Datasets.} Six real-world graph datasets with different homophilic extents are used and their statistics are listed in Table \ref{table:stat graph}.

\begin{table}[htbp]
  \centering
  \caption{\textbf{Statistics of six graph datasets.}}
  \resizebox{0.7\textwidth}{!}{%
  \label{table:stat graph}
  \begin{tabular}{cccccccccccccccccc}
  \toprule
  Dataset         & \texttt{CiteSeer} & \texttt{Cora}  & \texttt{PubMed}  & \texttt{Cornell} & \texttt{Texas}  & \texttt{Wisconsin} \\
  \midrule
  $|\mathcal{V}|$ & 3,327    & 2,708 & 19,717  & 183     &  183   & 251    \\
  $|\mathcal{E}|$ & 4,676    & 5,278 & 44,327  & 280     &  295   & 466    \\
  \#Features      & 3,703    & 1,433 & 500     & 1,703   &  1,703 & 1,703  \\
  \#Classes       & 7        & 6     & 3       & 5       &   5    & 5      \\
  Homophily Score             & 0.74     & 0.81  & 0.80    & 0.30    & 0.11   & 0.21   \\
  \bottomrule
  \end{tabular}%
  }
\end{table}

\textbf{Hypergraph Datasets.} Twelve benchmark hypergraph datasets are used in this work with diverse scales, structures, and homophilic extents. Their statistics are listed in Table \ref{table:stat hyper}.

\begin{table}[htbp]
  \centering
  \caption{\textbf{Statistics of twelve hypergraph datasets.}}
  \label{table:stat hyper}
  \resizebox{\textwidth}{!}{%
  \begin{tabular}{cccccccccccccc}
    \toprule
    Dataset         & \texttt{Cora} & \texttt{Citeseer} & \texttt{Pubmed} & \texttt{Cora-CA} & \texttt{DBLP-CA} & \texttt{ModelNet40} & \texttt{NTU2012} & \texttt{House} & \texttt{Zoo}   & \texttt{20News} & \texttt{Yelp}   & \texttt{Senate} \\
    \midrule
    $|\mathcal{V}|$ & 2,708 & 3,312     & 19,717  & 2,708    & 41,302   & 12,311      & 2,012    & 1,290  & 101   & 16,242  & 50,758  & 282 \\
    $|\mathcal{E}|$ & 1,579 & 1,079     & 7,963   & 1,072    & 22,363   & 12,311      & 2,012    & 341   & 42    & 100    & 679,302 & 315 \\
    \#Features      & 1,433 & 3,703     & 500    & 1,433    & 1,425    & 100        & 100     & 100   & 16    & 100    & 1,862   & 2 \\
    \#Classes        & 7    & 6        & 3      & 7       & 6       & 40         & 67      & 2     & 7     & 4      & 9      & 2 \\
    % $\max |e|$      & 5    & 26   & 171   & 43    & 202   & 93   & 2241 & 5    & 5     & 2838   & 81   & x \\
    Homophily Score    & 0.897&0.893     &0.952   &0.803    & 0.869   & 0.853      & 0.752   & 0.509 & 0.241 & 0.461  & 0.226  & 0.498 \\
    \bottomrule
  \end{tabular}
  }
\end{table}

\subsection{Baselines and Settings}

We compare our UniG-Encoder framework with several classic graph-oriented models, including (1) MLP; (2) general GNN methods: GCN~\cite{kipf2016semi}, GAT~\cite{velickovic2017graph}, GCNII~\cite{chen2020simple}, GraphSAGE~\cite{hamilton2017inductive}; (3) heterophily-oriented methods: H2GCN~\cite{zhu2020beyond}, GGCN~\cite{yan2022two}, GloGNN~\cite{li2022finding}, across various benchmark datasets. To conduct these experiments, we adopt ten random splits with a ratio of 48\%/32\%/20\% of nodes per class for training/validation/test, as previously established in~\cite{li2022finding}. We evaluate the performance by computing the overall mean accuracy and standard deviation on the test sets over the ten splits.

We also present a comparative analysis of our proposed framework UniG-Encoder against several state-of-the-art models on hypergraph benchmarks, including HGNN~\cite{feng2019hypergraph}, HCHA~\cite{bai2021hypergraph}, HNHN~\cite{dong2020hnhn}, HyperGCN~\cite{yadati2019hypergcn}, UniGCNII~\cite{huang2021unignn}, AllSet (AllDeepSets and AllSetTransformer)~\cite{chien2021you}, ED-HNN~\cite{wang2022equivariant}, and LE-GCN~\cite{yang2022semi}. To ensure a fair comparison, we follow the experimental protocols of ED-HNN for the hypergraph datasets experiments. Specifically, we split the data into training, validation, and test sets in a 50\%/25\%/25\% ratio, as suggested in~\cite{chien2021you}. We adopt prediction accuracy as evaluation metric and run each model ten times with different training and validation splits to obtain the mean accuracy and standard deviation.

\section{Results and Analysis}

\textbf{Overall Performance Analysis.} We present our experimental results on six graph datasets in Table \ref{table: acc-2}. It is noted that general GNN models such as GCN, GAT, GCNII, and GraphSAGE perform well on homophilic datasets such as \texttt{CiteSeer}, \texttt{Cora}, and \texttt{PubMed}, but their performance deteriorates on heterophilic datasets such as \texttt{Cornell}, \texttt{Texas}, and \texttt{Wisconsin}, even outperformed by simple models such as MLP. Our proposed framework not only achieves competitive performance compared to general GNNs, but also outperforms some heterophily-oriented models such as H2GCN, GGCN, and GloGNN, by adjusting the weights in the projection matrix $\mathbf{P}$. Details on this adjustment can be found in Appendix.

\begin{table}[htbp]
  \centering
  \caption{\textbf{Results on graphs.} Mean accuracy (\%) $\pm$ standard deviation is shown for each method. For each dataset, we mark the winner's score in bold and highlight the runner-up's with underline.}
  \label{table: acc-2}
  \resizebox{1.0\textwidth}{!}{%
  \begin{tabular}{ccccccc}
    \toprule
    Graph    & \texttt{CiteSeer}         & \texttt{Cora}             & \texttt{PubMed}           & \texttt{Cornell}   & \texttt{Texas}        & \texttt{Wisconsin} \\
    \midrule
    MLP       & $74.02 \pm 1.90$ & $75.69 \pm 2.00$ & $87.16 \pm 0.37$ & $81.89 \pm 6.40$ & $80.81 \pm 4.75$ & $85.29 \pm 3.31$ \\
    GCN       & $76.50 \pm 1.36$ & $86.98 \pm 1.27$ & $88.42 \pm 0.50$ & $60.54 \pm 5.30$ & $55.14 \pm 5.16$ & $51.76 \pm 3.06$ \\
    GCNII     & $\underline{77.33 \pm 1.48}$ & $\mathbf{88.37 \pm 1.25}$ & $\mathbf{90.15 \pm 0.43}$ & $77.86 \pm 3.79$ & $77.57 \pm 3.83$ & $80.39 \pm 3.40$ \\
    GraphSAGE & $76.04 \pm 1.30$ & $86.90 \pm 1.04$ & $88.45 \pm 0.50$ & $75.95 \pm 5.01$ & $82.43 \pm 6.14$ & $81.18 \pm 5.56$ \\
    GAT       & $76.55 \pm 1.23$ & $87.30 \pm 1.10$ & $86.33 \pm 0.48$ & $61.89 \pm 5.05$ & $52.16 \pm 6.63$ & $49.41 \pm 4.09$ \\
    % GCN-Cheby & $75.82 \pm 1.53$ & $86.76 \pm 0.95$ & $88.72 \pm 0.55$ & $74.32 \pm 7.46$ & $77.30 \pm 4.07$ & $79.41 \pm 4.46$ \\
    % GCN+JK    & $74.51 \pm 1.75$ & $85.79 \pm 0.92$ & $88.41 \pm 0.45$ & $64.59 \pm 8.68$ & $66.49 \pm 6.64$ & $74.31 \pm 6.43$ \\
    % Cheby+JK  & $74.98 \pm 1.18$ & $85.49 \pm 1.27$ & $89.07 \pm 0.30$ & $74.59 \pm 7.87$ & $78.38 \pm 6.37$ & $82.55 \pm 4.57$ \\
    H2GCN     & $77.07 \pm 1.64$ & $87.81 \pm 1.35$ & $89.59 \pm 0.33$ & $82.16 \pm 6.00$ & $82.16 \pm 5.28$ & $86.67 \pm 4.69$ \\
    GGCN      & $77.14 \pm 1.45$ & $87.95 \pm 1.05$ & $89.15 \pm 0.37$ & $85.68 \pm 6.63$ & $\underline{84.86 \pm 4.55}$ & $86.86 \pm 3.29$ \\
    GloGNN & $\mathbf{77.41 \pm 1.65}$ & $\underline{88.31 \pm 1.13}$ & $89.62 \pm 0.35$ & $\underline{85.95 \pm 5.10}$ & $84.32 \pm 4.15$ & $\underline{87.06 \pm 3.53}$ \\
    \midrule
    UniG-Encoder & $\underline{77.33 \pm 1.86}$  & $87.36 \pm 1.17$ & $\underline{89.76 \pm 0.46}$ & $\mathbf{86.75 \pm 6.56}$ & $\mathbf{85.40 \pm 5.3} $ & $\mathbf{88.03 \pm 4.42}$ \\
    \bottomrule
  \end{tabular}
  }
\end{table}

Table \ref{table: acc-1} illustrates the results of our comparative analysis, demonstrating that our proposed UniG-Encoder well performs on all twelve hypergraph datasets, compared to existing models, ranking 1st in 6/12 datasets and 2nd in 4/12 datasets. The top-performing baseline models include AllSetTransformer, ED-HNN, MLP, and LEGCN, etc. However, their performance varies significantly across different datasets. For instance, AllSetTransformer, AllDeepSets, UniGCNII, and ED-HNN exhibit promising results on homophilic hypergraph datasets such as citation networks, but their performance are subpar on heterophilic datasets, such as \texttt{House} and \texttt{Senate}, where MLP and LEGCN perform much better. In contrast, our framework achieves consistently superior results. The out-of-memory (OOM) issue in LE-GCN is caused by line expansion, which generates ``line nodes'' from hyperedges. In datasets such as \texttt{Yelp}, which contain a large number of hyperedges, OOM error may also occur due to memory constraint.

\begin{table}[htbp]
  \centering
  \caption{\textbf{Results on hypergraphs.} Mean accuracy (\%) $\pm$ standard deviation is shown for each method. For each dataset, we mark the winner's score in bold and highlight the runner-up's with underline. ``OOM'' denotes out-of-memory issue.}
  \label{table: acc-1}
  \resizebox{1.0\textwidth}{!}{%
  \begin{tabular}{ccccccc}
    \toprule
    Hypergraph & \texttt{Cora}             & \texttt{Citeseer}         & \texttt{Pubmed}           & \texttt{Cora-CA}          & \texttt{DBLP-CA}          & \texttt{ModelNet40}\\
    \midrule
    HGNN        & $79.39 \pm 1.36$ & $72.45 \pm 1.16$ & $86.44 \pm 0.44$ & $82.64 \pm 1.65$ & $91.03 \pm 0.20$ & $95.44 \pm 0.33$  \\
    HCHA        & $79.14 \pm 1.02$ & $72.42 \pm 1.42$ & $86.41 \pm 0.36$ & $82.55 \pm 0.97$ & $90.92 \pm 0.22$ & $94.48 \pm 0.28$  \\
    HNHN        & $76.36 \pm 1.92$ & $72.64 \pm 1.57$ & $86.90 \pm 0.30$ & $77.19 \pm 1.49$ & $86.78 \pm 0.29$ & $97.84 \pm 0.25$  \\
    HyperGCN    & $78.45 \pm 1.26$ & $71.28 \pm 0.82$ & $82.84 \pm 8.67$ & $79.48 \pm 2.08$ & $89.38 \pm 0.25$ & $75.89 \pm 5.26$  \\
    UniGCNII    & $78.81 \pm 1.05$ & $73.05 \pm 2.21$ & $88.25 \pm 0.40$ & $83.60 \pm 1.14$ & $\underline{91.69 \pm 0.19}$ & $98.07 \pm 0.23$  \\
    AllDeepSets & $76.88 \pm 1.80$ & $70.83 \pm 1.63$ & $88.75 \pm 0.33$ & $81.97 \pm 1.50$ & $91.27 \pm 0.27$ & $96.98 \pm 0.26$  \\
    AllSetTransformer  & $78.58 \pm 1.47$ & $73.08 \pm 1.20$ & $88.72 \pm 0.37$ & $83.63 \pm 1.47$ & $91.53 \pm 0.23$ & $98.20 \pm 0.20$   \\
    ED-HNN & $\underline{80.31 \pm 1.35}$ & $73.70 \pm 1.38$ & $\mathbf{89.56 \pm 0.62}$ & $\underline{83.97 \pm 1.55}$ & $\mathbf{91.93 \pm 0.29}$ & $\underline{98.35 \pm 0.20}$  \\
    LE-GCN & $77.34 \pm 1.10$ & $73.41 \pm 1.15$ & $88.53 \pm 0.48$ & $76.60 \pm 1.63$ & $85.82 \pm 0.31$ & $96.68 \pm 0.16$  \\
    MLP & $77.49 \pm 1.43 $ & $\underline{73.99 \pm 0.85} $  & $88.50 \pm 0.39$ & $77.40 \pm 1.38$ & $85.85 \pm 0.43$ & $96.70 \pm 0.23$ \\
    \midrule
    UniG-Encoder & $\mathbf{81.43 \pm 1.37} $ & $\mathbf{75.08 \pm 1.45}$  & $\underline{88.98 \pm 0.37}$ & $\mathbf{85.58 \pm 1.13}$ & $91.65 \pm 0.15$ & $\mathbf{98.41 \pm 0.17}$  \\
    \bottomrule
    \toprule
    Hypergraph & \texttt{NTU2012}          & \texttt{Zoo}       & \texttt{20Newsgroups}     & \texttt{Yelp}             & \texttt{House}            & \texttt{Senate} \\
    \midrule
    HGNN        & $87.72 \pm 1.35$ & $95.50 \pm 4.58$ & $80.33 \pm 0.42$ & $33.04 \pm 0.62$ & $61.39 \pm 2.96$ & $48.59 \pm 4.52$ \\
    HCHA        & $87.48 \pm 1.87$ & $93.65 \pm 6.15$ & $80.33 \pm 0.80$ &  $30.99 \pm 0.72$ & $61.36 \pm 2.53$ & $48.62 \pm 4.41$ \\
    HNHN        & $89.11 \pm 1.44$ & $93.59 \pm 5.88$ & $81.35 \pm 0.61$ & $31.65 \pm 0.44$ & $67.80 \pm 2.59$ & $50.93 \pm 6.33$ \\
    HyperGCN    & $56.36 \pm 4.86$ & $85.38 \pm 6.23$ & $81.05 \pm 0.59$ & $29.42 \pm 1.54$ & $48.32 \pm 2.93$ & $42.45 \pm 3.67$ \\
    UniGCNII    & $\underline{89.30 \pm 1.33}$ & $93.65 \pm 4.37$ & $81.12 \pm 0.67$ & $31.70 \pm 0.52$ & $67.25 \pm 2.57$ & $49.30 \pm 4.25$ \\
    AllDeepSets & $88.09 \pm 1.52$ & $95.39 \pm 4.77$ & $81.06 \pm 0.54$ & $30.36 \pm 1.57$ & $67.82 \pm 2.40$ & $48.17 \pm 5.67$ \\
    AllSetTransformer & $88.69 \pm 1.24$ & $\underline{97.50 \pm 3.59}$ & $81.38 \pm 0.58$ & $\mathbf{36.89 \pm 0.51}$ & $69.33 \pm 2.20$ & $51.83 \pm 5.22$  \\
    ED-HNN & $88.07 \pm 1.28$ & $95.77 \pm 3.37$ & $\mathbf{81.90 \pm 0.55}$ & $34.99 \pm 0.55$ & $72.45 \pm 2.28$ & $64.79 \pm 5.14$ \\
    LE-GCN  & $89.16 \pm 1.13$ & $95.00 \pm 4.81$ & $\underline{81.84 \pm 0.34}$ & OOM & $78.39 \pm 1.64$ & $\mathbf{80.70 \pm 5.67}$ \\
    MLP     & $89.08 \pm 1.58$  & $94.62 \pm 4.51$ & $81.42 \pm 0.49 $ & $32.67 \pm 0.32$ & $\mathbf{78.79 \pm 2.28}$ & $79.72 \pm 3.40$ \\
    \midrule
    UniG-Encoder & $\mathbf{90.42 \pm 1.49}$  & $\mathbf{98.46 \pm 3.71}$ & $81.74 \pm 0.54$ & $\underline{36.33 \pm 0.28}$ & $\underline{78.73 \pm 2.00}$ & $\underline{80.56 \pm 3.86}$ \\
    \bottomrule
  \end{tabular}
  }
\end{table}

\textbf{Impacts of Normalization.} We compare five types of normalization here, including (1) no normalization for $\mathbf{P}$ and $\mathbf{P}^{\top}$, (2) row normalization for $\mathbf{P}$ and $\mathbf{P}^{\top}$, (3) column normalization for $\mathbf{P}$ and $\mathbf{P}^{\top}$, (4) row normalization for $\mathbf{P}$ and column normalization for $\mathbf{P}^\top$, and (5) column normalization for $\mathbf{P}$ and row normalization for $\mathbf{P}^{\top}$. We conduct experiments on the \texttt{Cora} hypergraph dataset. It is indicated that normalizing $\mathbf{P}$ and $\mathbf{P}^{\top}$ by row produces the best results ($79.01\pm1.40$, $80.64\pm1.62$, $80.22\pm0.76$, $80.03\pm0.97$, $80.25\pm1.17$ for types (1)-(5) respectively). This finding is also consistent with our design of $\mathbf{P}_\mathcal{E}$, which represents a weighted average on edges/hyperedges, while $\mathbf{P}^{\top}$ is used to aggregate the embeddings of the nodes and their neighbors.

\begin{figure}[htbp]
  \centering
  % \subfigure[CE-Based Expansion Homophily]{
  \subfigure[Real-World Datasets]{
      \includegraphics[width=0.43\textwidth]{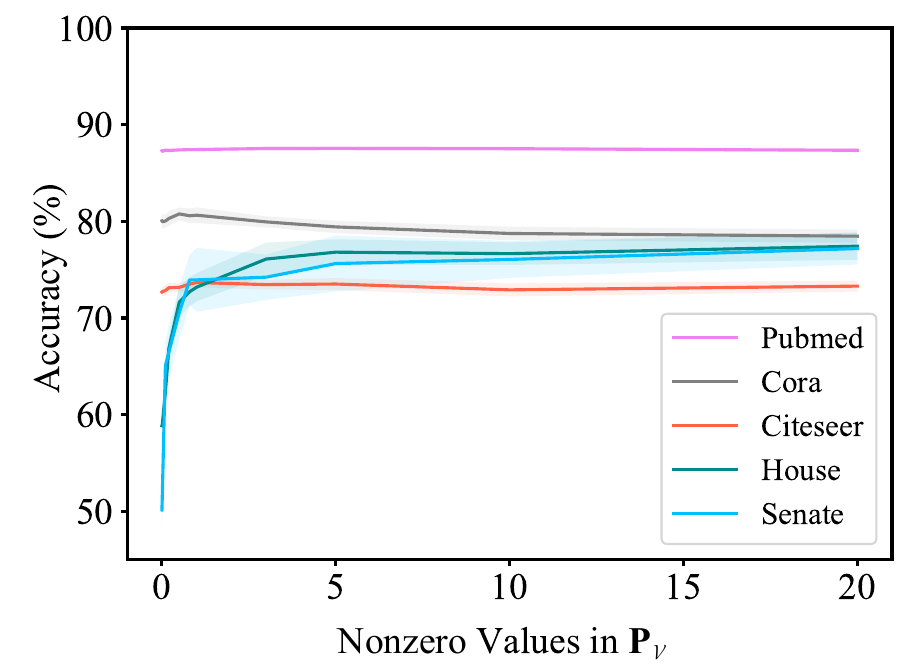}
  }\hspace{8mm}
  \subfigure[Synthetic \texttt{Texas} Datasets]{
      \includegraphics[width=0.43\textwidth]{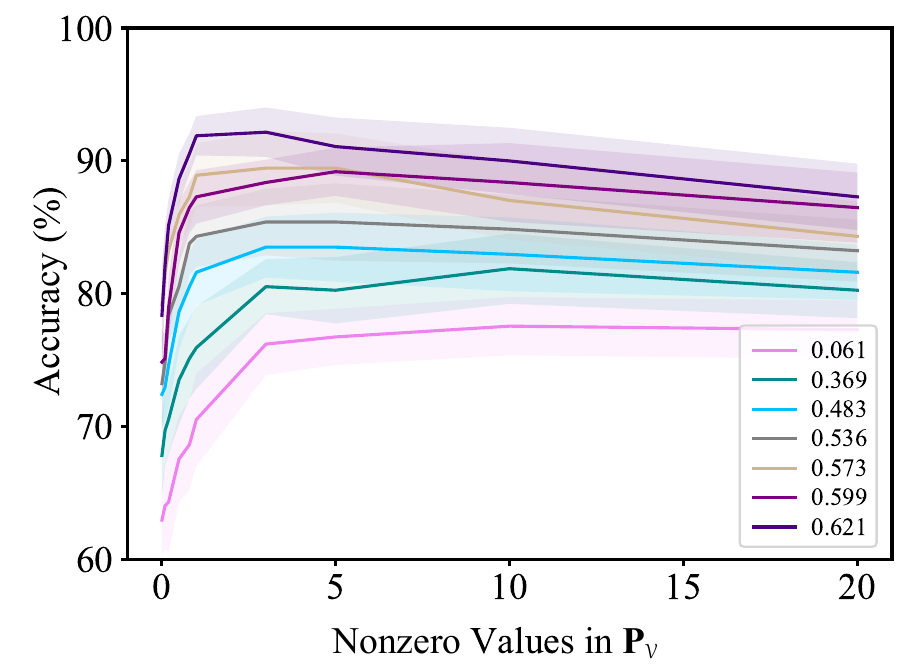}
  }
  % \subfigure[PubMed \protect\newline (GCN, 88.42±0.50)]{
  %     \includegraphics[width=0.3\textwidth]{PubMed--Neighborhood Similarity.pdf}
  % }
  \caption{\textbf{Performance with different nonzero values in $\mathbf{P}_\mathcal{V}$.} The legends in (b) correspond to the Homophily Scores of the synthetic datasets.}
  \label{fig:q1}
\end{figure}

\textbf{Impacts of weight on $\mathbf{P}_\mathcal{V}$.} Our UniG-Encoder enables tuning the weight on $\mathbf{P}_\mathcal{V}$ to accommodate various scenarios, especially with different homophilic extents. Heterophily refers to a situation where a node's neighbors are substantially different from the node itself. In this case, when performing operations such as summation or averaging on the features of a node and its neighbors, the \textit{ego-feature} should dominate. In our approach, this can be simply realized by modifying the nonzero values in $\mathbf{P}_\mathcal{V}$ to control the weights of features from the node itself and its neighbors. This technique also enables a performance enhancement of our framework on heterophilic datasets. To validate, we conduct experiments on both real-world datasets and synthetic datasets obtained from \texttt{Texas}. The results are depicted in Figure \ref{fig:q1}. As illustrated in Figure \ref{fig:q1}(a), for datasets \texttt{Pubmed}, \texttt{Cora}, and \texttt{Citeseer} with high degree of homophily, the variation in the nonzero values of $\mathbf{P}_\mathcal{V}$ has minor impact on their performance in the node classification task. This result aligns with our intuition that when a node and its neighbors have consistent features, the contribution of neighbors to the aggregated embeddings should be similar to that of the node itself. However, in a heterophilic graph where nodes and their neighbors have different categories, achieving better classification performance requires a trade-off between \textit{ego-features} and the features of neighbors. Consequently, excessively small or large nonzero values in $\mathbf{P}_\mathcal{V}$ will both result in suboptimal accuracy, as depicted in Figure \ref{fig:q1}(b).

\textbf{Impacts of Projection Placement.} In fact, the projection operation with its reverse in our framework can be placed at any layer of the neural network pipeline. We present experiments placing projection and its reverse (denoted by ``$\cdot~\&~\cdot$'') at different layers of a three-layer MLP, and the results are shown in Table \ref{table: ab-3}. Although the effect of placing projection at different layers is not that significant, we emphasize that two variants, $\mathbf{PX}\&\mathbf{P^{\top}H^{(0)}}$ and $\mathbf{PH^{(1)}}\&\mathbf{P^{\top}H^{(2)}}$, which in fact successively execute forward projection and its reverse by multiplying the embeddings by $\mathbf{P}^\top\mathbf{P}$, which can be regarded as a decomposition of the adjacency matrix~\cite{yang2015network,huang2017accelerated}, practically do not increase the time complexity. It is worth noting that the framework can be easily extended to utilize multi-hop neighborhood information by using multiple $\mathbf{P}^\top\mathbf{P}$.

\begin{table}[htbp]
  \centering
  \caption{\textbf{Results of different projection placement.}}
  \label{table: ab-3}
  \resizebox{0.8\textwidth}{!}{%
  \begin{tabular}{ccccc}
    \toprule
    Variant    &  \texttt{Cora}   & \texttt{Citeseer}  &  \texttt{Pubmed}  & \texttt{House} \\
    \midrule
    No Projection & $75.47 \pm 1.27$ & $73.47 \pm 1.05$ & $88.41 \pm 0.54$ & $76.93 \pm 2.67$ \\
    $\mathbf{PX}$ \& $\mathbf{P^{\top}H^{(0)}}$ & $79.97 \pm 1.14$ & $74.12 \pm 1.09$ & $ 88.68 \pm 0.48 $ & $77.00 \pm 2.99$ \\
    $\mathbf{PX}$ \& $\mathbf{P^{\top}H^{(1)}}$ & $80.34 \pm 1.22$ & $74.35 \pm 1.09$ & $ 88.70 \pm 0.50 $ & $76.44 \pm 3.46$ \\
    $\mathbf{PX}$ \& $\mathbf{P^{\top}H^{(2)}}$ & $80.31 \pm 1.17$ & $74.36 \pm 1.09$ & $ 88.73 \pm 0.41 $ & $76.44 \pm 3.46$\\
    $\mathbf{PH^{(0)}}$ \& $\mathbf{P^{\top}H^{(1)}}$ & $80.64 \pm 1.19$ & $74.44 \pm 1.25$ & $88.60 \pm 0.42$ & $77.21 \pm 2.68$\\
    $\mathbf{PH^{(0)}}$ \& $\mathbf{P^{\top}H^{(2)}}$ & $80.66 \pm 1.21$ & $74.43 \pm 1.24$ & $88.66 \pm 0.61$ & $ 77.24 \pm 2.68$ \\
    $\mathbf{PH^{(1)}}$ \& $\mathbf{P^{\top}H^{(2)}}$ & $80.44 \pm 1.00$ & $74.44 \pm 1.25$ & $88.60 \pm 0.42$ & $77.24 \pm 2.68$ \\
    % Proj-GNN & $79.01 \pm 1.40$  & $80.03 \pm 0.97$ \\
    \bottomrule
  \end{tabular}
  }
\end{table}

\textbf{Complexity Analysis.} Generally, the proposed framework has a similar computing complexity as the used neural network, such as MLP, Transformer. The extra computing consumption is brought in by the dimension increase between the forward projection and its reverse. Therefore, as mentioned above, if we place the forward projection and its reverse adjacently, i.e., multiplying the embeddings directly by $\mathbf{P}^\top\mathbf{P}$, no extra complexity exists.

\section{Conclusion}

In this study, we propose a new universal architecture for both graph and hypergraph representation learning, called UniG-Encoder. In contrast to the traditional spectral-based and/or message passing approaches, our proposed framework simultaneously and comprehensively exploits the node features and graph/hypergraph topologies in an efficient and unified manner, covering both heterophilic and homophilic graphs. The designed projection matrix, serving as key encoder to the graph features, is intuitive and interpretable. We conduct experiments on various graph and hypergraph datasets with different scales, structures, and homophilic extents. The experimental results and comparisons with the state-of-the-art methods demonstrate superior performance of the proposed UniG-Encoder. The framework can lead to potential applications in many tasks, such as graph classification and link prediction.

\appendix
\section{Proofs}

\subsection{Proof of Theorem 1}

\textit{Proof.} Let $\sigma:\mathcal{V}\to\mathcal{V}$ be an arbitrary permutation. The map $\phi$ can be constructed as follows: for $v\in\mathcal{V}$, let $\phi(v)=\sigma(v)$; for $e\in\mathcal{E}$, let $\phi(e)=e$. As there is no duplicate edge/hyperedge in $\mathcal{G}$, it is clear that $\phi|_\mathcal{V}=\sigma$ and $\phi_\mathcal{E}$ is an identity map on $\mathcal{E}$. Because $\sigma$ and the identity map are bijective and $\mathcal{V}\cap\mathcal{E}=\emptyset$, the map $\phi$ is also bijective.

\subsection{Proof of Theorem 2}

\textit{Proof.} Note that $\mathbf{P}_{\mathcal{V},ij}^\top=\delta_{i,\sigma(j)}$. Thus $(\mathbf{P}_\mathcal{V}^\top\mathbf{P}_\mathcal{V})_{ij}=\sum_{k=1}^{|\mathcal{V}|}\mathbf{P}_{\mathcal{V},ik}^\top\mathbf{P}_{\mathcal{V},kj}=\sum_{k=1}^{|\mathcal{V}|}\delta_{i,\sigma(k)}\delta_{j,\sigma(k)}=\delta_{ij}$, which means that $\mathbf{P}_\mathcal{V}^\top\mathbf{P}_\mathcal{V}=\mathbf{I}$. Note also that $\mathbf{P}_{\mathcal{E},ij}$ equals to $1$ if $v_j\in e_i$. Thus $\mathbf{P}_{\mathcal{E},ij}^\top=1$ if $v_i\in e_j$. It follows that $(\mathbf{P}_\mathcal{E}^\top\mathbf{P}_\mathcal{E})_{ij}=\sum_{k=1}^{|\mathcal{V}_\mathcal{E}|}\mathbf{P}_{\mathcal{E},ik}^\top\mathbf{P}_{\mathcal{E},kj}=\sum_{k=1}^{|\mathcal{V}_\mathcal{E}|}1_{\{v_i\in e_k\}}1_{\{v_j\in e_k\}}$, where $1_{\{\cdot\}}$ is the indicator function. Therefore, $\mathbf{P}^\top\mathbf{P}=\left[\mathbf{P}^\top_\mathcal{V},\mathbf{P}^\top_\mathcal{E}\right]\left[\begin{matrix}\mathbf{P}_\mathcal{V}\\\mathbf{P}_\mathcal{E}\end{matrix}\right]=\mathbf{P}_\mathcal{V}^\top\mathbf{P}_\mathcal{V}+\mathbf{P}_\mathcal{E}^\top\mathbf{P}_\mathcal{E}=\mathbf{I}+\mathbf{P}_\mathcal{E}^\top\mathbf{P}_\mathcal{E}$. This indicates that $(\mathbf{P}^\top\mathbf{P})_{ij}>0$ if and only if $\mathbf{I}_{ij}>0$ or $(\mathbf{P}_\mathcal{E}^\top\mathbf{P}_\mathcal{E})_{ij}>0$, which proves the theorem.

\section{A Schematic Example}

We provide here a schematic example to show the intuition and interpretability of the projection matrix. For the hypergraph shown in Figure 1 of the main text, the incidence matrix
$$
\mathbf{B}={\left[\begin{array}{l l l}
  {1}&{0}&{0}\\
  {1}&{0}&{0}\\
  {1}&{1}&{0}\\
  {0}&{1}&{0}\\
  {1}&{0}&{1}\\
  {0}&{0}&{1}\\
  {0}&{0}&{1}\\
\end{array}\right]}.
$$
Thus the projection matrix without permutation on $\mathbf{P}_\mathcal{V}$ is
$$
\mathbf{P}=\left[\begin{array}{lllllll}
1 & 0 & 0 & 0 & 0 & 0 & 0 \\
0 & 1 & 0 & 0 & 0 & 0 & 0 \\
0 & 0 & 1 & 0 & 0 & 0 & 0 \\
0 & 0 & 0 & 1 & 0 & 0 & 0 \\
0 & 0 & 0 & 0 & 1 & 0 & 0 \\
0 & 0 & 0 & 0 & 0 & 1 & 0 \\
0 & 0 & 0 & 0 & 0 & 0 & 1 \\
1 & 1 & 1 & 0 & 1 & 0 & 0 \\
0 & 0 & 1 & 1 & 0 & 0 & 0 \\
0 & 0 & 0 & 0 & 1 & 1 & 1
\end{array}\right]
$$
and its transpose is
$$
\mathbf{P}^{\top}=\left[\begin{array}{llllllllll}
1 & 0 & 0 & 0 & 0 & 0 & 0 & 1 & 0 & 0\\
0 & 1 & 0 & 0 & 0 & 0 & 0 & 1 & 0 & 0\\
0 & 0 & 1 & 0 & 0 & 0 & 0 & 1 & 1 & 0\\
0 & 0 & 0 & 1 & 0 & 0 & 0 & 0 & 1 & 0\\
0 & 0 & 0 & 0 & 1 & 0 & 0 & 1 & 0 & 1\\
0 & 0 & 0 & 0 & 0 & 1 & 0 & 0 & 0 & 1\\
0 & 0 & 0 & 0 & 0 & 0 & 1 & 0 & 0 & 1\\
\end{array}\right].
$$
The compound matrix thus satisfies
$$
\mathbf{P}^{\top}\mathbf{P}=[\mathbf{I},\mathbf{B}]\left[\begin{matrix}\mathbf{I}\\\mathbf{B}^\top\end{matrix}\right]=\mathbf{I}+\mathbf{B}\mathbf{B}^\top=\left[\begin{array}{lllllll}
  2& 1& 1& 0& 1& 0& 0 \\
  1& 2& 1& 0& 1& 0& 0 \\
  1& 1& 3& 1& 1& 0& 0 \\
  0& 0& 1& 2& 0& 0& 0 \\
  1& 1& 1& 0& 3& 1& 1 \\
  0& 0& 0& 0& 1& 2& 1 \\
  0& 0& 0& 0& 1& 1& 2 \\
\end{array}\right].
$$
Note that the adjacency matrix $\mathbf{A}$ of a graph or hypergraph is defined as $\mathbf{B}\mathbf{B}^\top$, where $\mathbf{A}_{ij}$ represents the number of edges/hyperedges shared between nodes $v_i$ and $v_j$. Therefore, $\mathbf{P}^{\top}\mathbf{P}=\mathbf{I}+\mathbf{A}$.

Regarding to the normalization process, the compound row normalized matrix is
$$
\begin{aligned}\hat{\mathbf{P}}^{\top}\hat{\mathbf{P}}&=\left[\mathbf{I}+\mathbf{D}_\mathcal{V}\right]^{-1}\left[\mathbf{I},\mathbf{B}\right]\left[\begin{array}{cc}{\mathbf{I}}&\mathbf{0}\\\mathbf{0}&{\mathbf{D}_\mathcal{E}}^{-1}\end{array}\right]{\left[\begin{matrix}{\mathbf{I}}\\{\mathbf{B}^{\top}}\end{matrix}\right]}\\
&=\left[\mathbf{I}+\mathbf{D}_\mathcal{V}\right]^{-1}\left[\mathbf{I},\mathbf{B}\right]\left[\begin{array}{c}{\mathbf{I}}\\{\mathbf{D}_\mathcal{E}}^{-1}\mathbf{B}^{\top}\end{array}\right]\\
&=\left[\mathbf{I}+\mathbf{D}_\mathcal{V}\right]^{-1}\left[\mathbf{I}+\mathbf{B}{\mathbf{D}_\mathcal{E}}^{-1}\mathbf{B}^{\top}\right],\end{aligned}
$$
where $\left[\mathbf{I}+\mathbf{D}_\mathcal{V}\right]^{-1}$ represents the row normalization factor for $\mathbf{P}^{\top}$ and $\left[\begin{array}{cc}{\mathbf{I}}&\mathbf{0}\\\mathbf{0}&{\mathbf{D}_\mathcal{E}}^{-1}\end{array}\right]$ for $\mathbf{P}$.

\section{Graph and Hypergraph Datasets}

We utilize a total of 6 representative graph datasets and 12 benchmark hypergraph datasets sourced from the existing literatures, with their statistics listed in Table 1 and 2 of the main text. Here we describe the detailed information of these datasets.

The graph datasets include \texttt{Cora}, \texttt{Citeseer}, \texttt{Pubmed}, \texttt{Texas}, \texttt{Wisconsin}, and \texttt{Cornell}~\cite{pei2020geom}. The \texttt{Cora}, \texttt{Citeseer}, and \texttt{Pubmed} datasets consist of citation graphs where nodes represent papers and edges denote the citation or quotation relationships between them. These graphs employ bag-of-words representations as the feature vectors for the nodes, indicating the presence of corresponding words from the dictionary in the papers. The labels in these datasets correspond to the classes or fields of the papers. The \texttt{Texas}, \texttt{Wisconsin}, and \texttt{Cornell} datasets comprise web pages collected from the computer science departments of their respective universities. In these datasets, nodes represent web pages, while edges represent hyperlinks connecting them. Each page in these datasets also employs bag-of-words representations as the feature vectors for the nodes, indicating the existence of corresponding words from the dictionary.

The benchmark hypergraph datasets include \texttt{Cora}, \texttt{Citeseer}, \texttt{Pubmed}, \texttt{Cora-CA}, \texttt{DBLP-CA}, \texttt{20Newsgroups}, \texttt{Zoo}, \texttt{ModelNet40}, \texttt{NTU2012}, \texttt{Yelp}, \texttt{House}, and \texttt{Senate}. The co-citation networks \texttt{Cora}, \texttt{Citeseer}, and \texttt{Pubmed}, are obtained from~\cite{yadati2019hypergcn}, in which all documents cited by a document are connected by a hyperedge. The co-authorship networks \texttt{Cora-CA} and \texttt{DBLP-CA} are also obtained from~\cite{yadati2019hypergcn}, in which all documents co-authored by an author are connected by a hyperedge. In these co-citation and co-authorship networks datasets, the node features consist of bag-of-words representations of the corresponding documents, and node labels are the paper classes. The \texttt{20Newsgroups} and \texttt{Zoo} datasets are obtained from the UCI Categorical Machine Learning Repository~\cite{asuncion2007uci}. In the \texttt{20Newsgroups} dataset, the node features consist of TF-IDF representations of news messages. In the \texttt{Zoo} dataset, the node features are combinations of categorical and numerical measurements describing various animals. Two public 3D object datasets in computer vision, namely \texttt{ModelNet40}~\cite{wu20153d} and \texttt{NTU2012}~\cite{chen2003visual}, are utilized. The former comprises of 12,311 3D objects from 40 categories, while the latter consists of 2,012 3D shapes from 67 categories. The two datasets feature visual objects with extracted features using the Group-View Convolutional Neural Network (GVCNN)~\cite{feng2018gvcnn} and the Multi-View Convolutional Neural Network (MVCNN)~\cite{su2015multi}. The construction of the hypergraphs follows the methodology described in~\cite{feng2019hypergraph,yang2022semi}. The \texttt{Yelp}, \texttt{House}, and \texttt{Senate} datasets are introduced in~\cite{chien2021you,fowler2006legislative}. Using the ``restaurant'' catalog in \texttt{Yelp}, all businesses are selected as nodes, and hyperedges are formed by selecting restaurants visited by the same user. The node labels, representing the average review of a restaurant, are derived from the numbers of rating stars, ranged from 1 to 5 stars with an interval of 0.5 star. The node features are constructed using the latitude, longitude, city, state (encoded as one-hot vectors), and bag-of-words encodings of the top-1000 words in the restaurant names. In the \texttt{House} dataset, each node represents a member of the US House of Representatives, and hyperedges are formed by grouping together members of the same committee. The node labels indicate the political party affiliation of the representatives. As the original \texttt{House} dataset lacks node features, they are generated using Gaussian random vectors, following a similar approach of the contextual stochastic block model. The feature vectors are fixed at a dimension of 100, and the features are obtained by applying one-hot encodings to the labels, with Gaussian noise $\mathcal{N}(\mathbf{0},\sigma^2\mathbf{I})$ added. The standard deviation of the noise, $\sigma$, is set to 1 here. In \texttt{Senate} dataset, nodes are US Congressperson and hyperedges are comprised of the sponsors and co-sponsors of bills put forth in the Senate. Each node in the datasets is labeled with political party affiliation.

\section{Experimental Settings}

All the experiments are conducted on a Linux machine running Ubuntu 18.04, equipped with eight NVIDIA 3090ti GPUs with 24GB memory. To ensure a fair comparison, we follow the same training recipe as~\cite{wang2022equivariant,chien2021you}. Adam optimizer~\cite{kingma2014adam} with fixed learning rate and weight decay across epochs is utilized to minimize the cross-entropy loss function. The models are trained for 500 epochs for all datasets. Dropout is applied to prevent overfitting, and \texttt{ReLU} is chosen as the nonlinear activation function. The best hyperparameters are determined using Optuna~\cite{akiba2019optuna} with 200 trails. The search range for the number of layers is $\{1,2\}$, and the hidden dimensions are selected from $\{64,128,256,512\}$. We tune the learning rate from the set $\{0.1,0.02,0.01,0.001,0.0001\}$, the weight decay from $\{0,0.005,0.0005,0.00005\}$, and the dropout rate from $\{0,0.5,0.7,0.9\}$. The initial nonzero values of $\mathbf{P}_\mathcal{V}$ are set to either $\{1,10,100,0.1,0.001,0.0001\}$ or $d_i\times\{1,10,100,0.1,0.001,0.0001\}$, where $d_i$ is the $i$-th diagonal value of $\mathbf{D}_\mathcal{V}$. $\mathbf{P}$ and $\mathbf{P}^\top$ are row normalized. The reported standard deviations are calculated by conducting experiments on ten different data splits.

\section{Sensitivity to Hidden Dimension}

We perform comparison study on the expressivity of UniG-Encoder versus the hidden dimension of the MLP, as shown in Table \ref{table: ab-5}, where we use different hidden dimensions and evaluate the performance on the \texttt{Pubmed} hypergraph dataset. We also compare our framework with the top-performing baselines, namely AllDeepSets, AllSetTransformer, and ED-HNN. Remarkably, our model with a hidden dimension of 128 achieves comparable results to the 512-width AllSet models and shows performance on par with the ED-HNN model. These results indicate that our UniG-Encoder exhibits good tolerance for low hidden dimension, which can be attributed to its enhanced expressive power via the normalized projection matrix.

\begin{table}[htbp]
  \centering
  \caption{\textbf{Sensitivity to hidden dimension on \texttt{Pubmed} hypergraph dataset.}}
  \label{table: ab-5}
  \resizebox{0.7\textwidth}{!}{%
  \begin{tabular}{ccccc}
    \toprule
    Model    &  512   & 256  &  128  & 64 \\
    \midrule
    % mlp (w/o) & $75.47 \pm 1.27$ & $73.47 \pm 1.05$ & $88.41 \pm 0.54$ & $76.93 \pm 2.67$ \\
    % $\mathbf{PX}$ + $\mathbf{P^{\top}H^{(0)}}$ & $79.97 \pm 1.14$ & $74.12 \pm 1.09$ & $ 88.68 \pm 0.48 $ & $77.00 \pm 2.99$ \\
    AllDeepSets & $88.75 \pm 0.33$ & $88.41 \pm 0.37$ & $87.50 \pm 0.42$ & $86.78 \pm 0.40$ \\
    AllSetTransformer & $88.72 \pm 0.37$ & $88.16 \pm 0.24$ & $87.36 \pm 0.23$ & $86.21 \pm 0.25$ \\
    ED-HNN & $89.03 \pm 0.53$ & $88.74 \pm 0.38$ & $88.84 \pm 0.38$ & $88.76 \pm 0.24$ \\
    UniG-Encoder & $88.98 \pm 0.37$ & $88.82 \pm 0.40$ & $ 88.83 \pm 0.48$ & $88.46 \pm 0.23$ \\
    % Proj-GNN & $79.01 \pm 1.40$  & $80.03 \pm 0.97$ \\
    \bottomrule
  \end{tabular}
  }
\end{table}

\section{Over-Smoothing Analysis}

\begin{figure}[htbp]
  \centering
  \subfigure[\texttt{Cora}]{
      \includegraphics[width=0.3\textwidth]{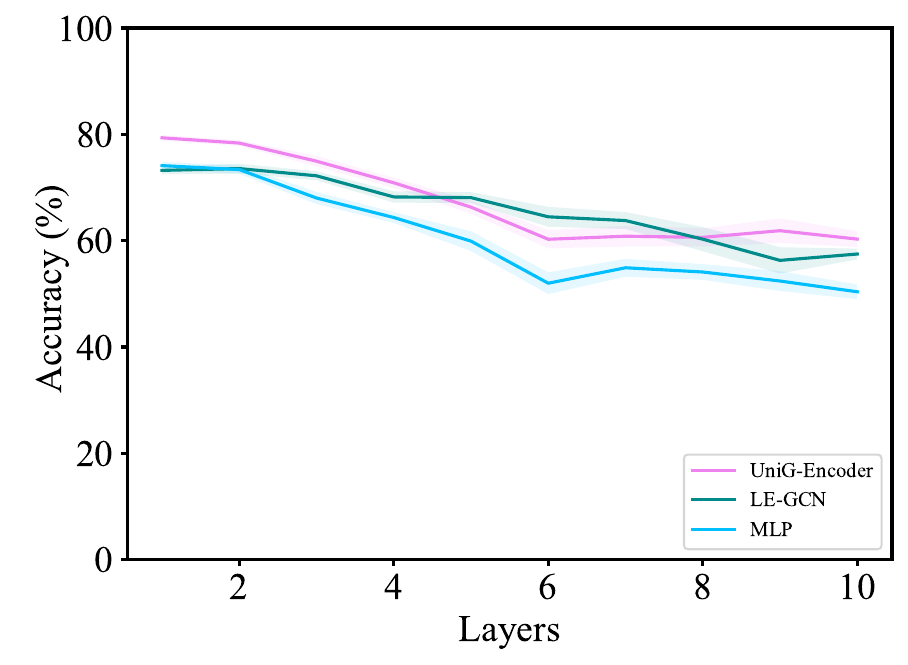}
  }
  \subfigure[\texttt{Citeseer}]{
      \includegraphics[width=0.3\textwidth]{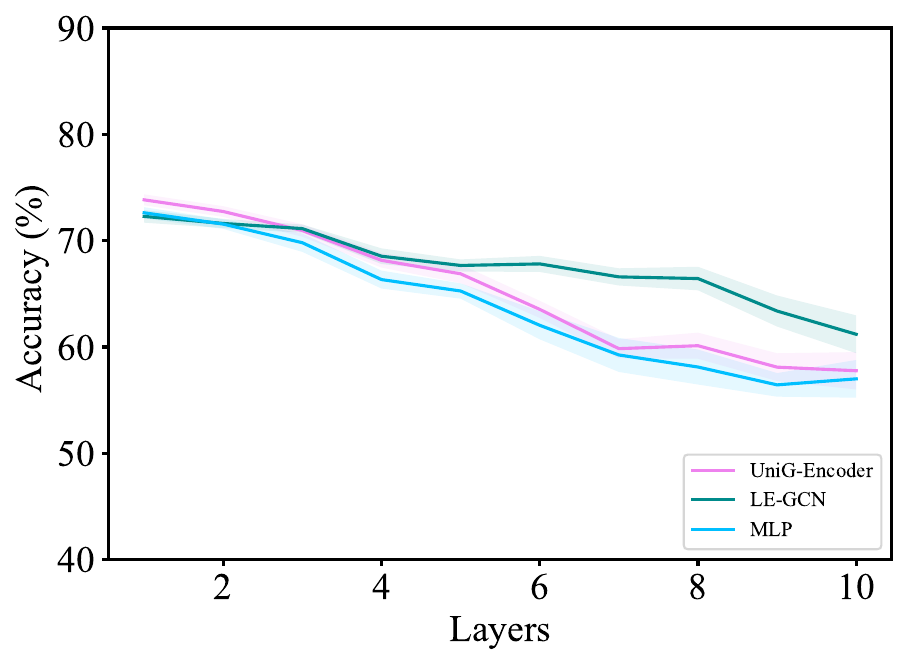}
  }
  \subfigure[\texttt{Pubmed}]{
    \includegraphics[width=0.3\textwidth]{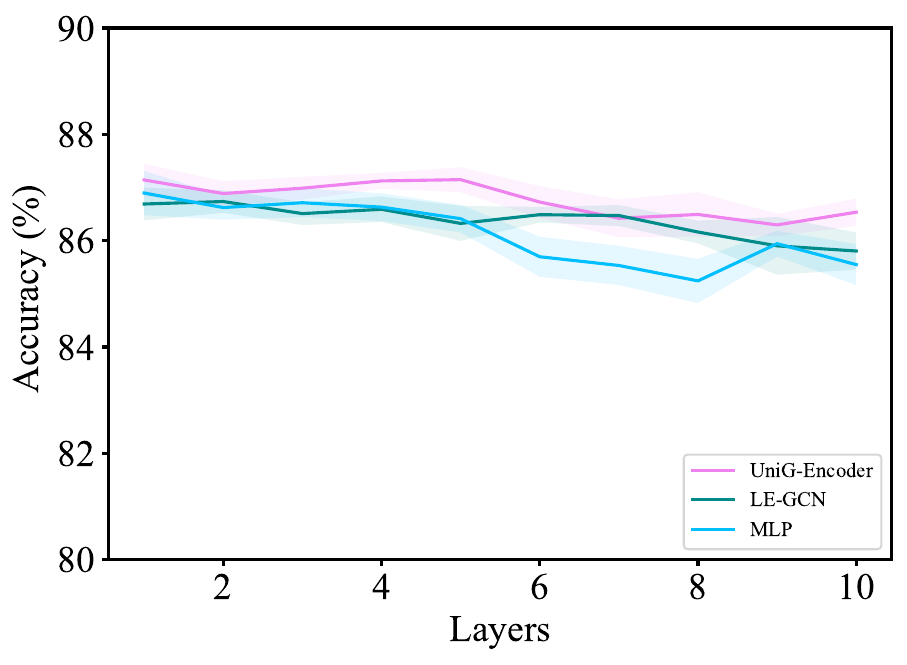}
}
  \caption{\textbf{Performance of different models with deeper layers on three benchmark datasets.}}
  \label{fig:q8-9}
\end{figure}

GNNs encounter over-smoothing problem when they are extended to deeper architectures. The mixing of node embeddings from different classes results in a decline in GNNs performance, due to the excessive aggregation of neighborhood information. Figure \ref{fig:q8-9} illustrates that as the models go deeper, their overall performance deteriorates due to over-smoothing.

\section{Synthetic Graph and Hypergraph Datasets}

\begin{figure}[htbp]
  \centering
  \subfigure[]{
      \includegraphics[width=0.3\textwidth]{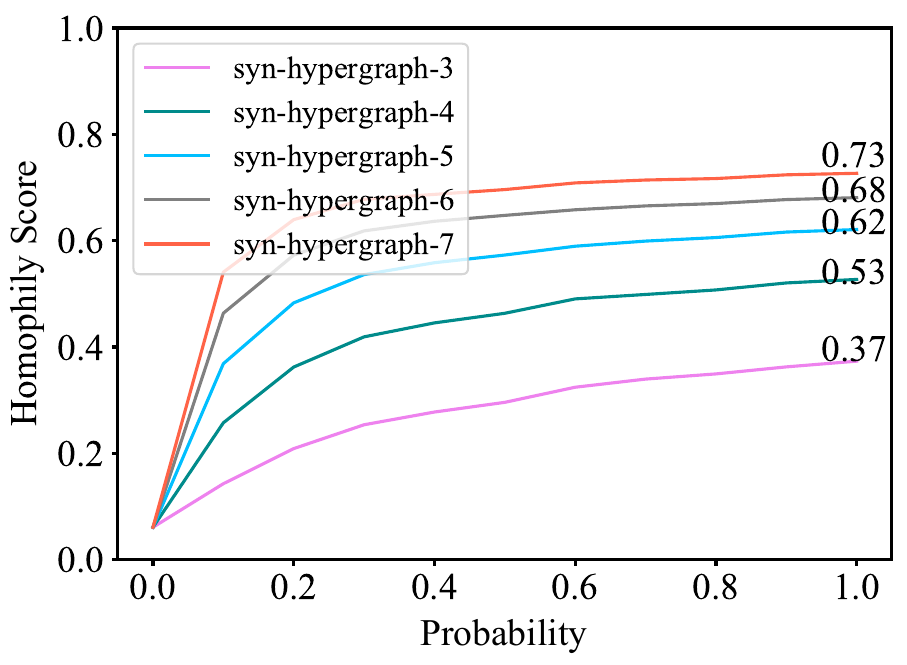}
  }
  \subfigure[]{
      \includegraphics[width=0.3\textwidth]{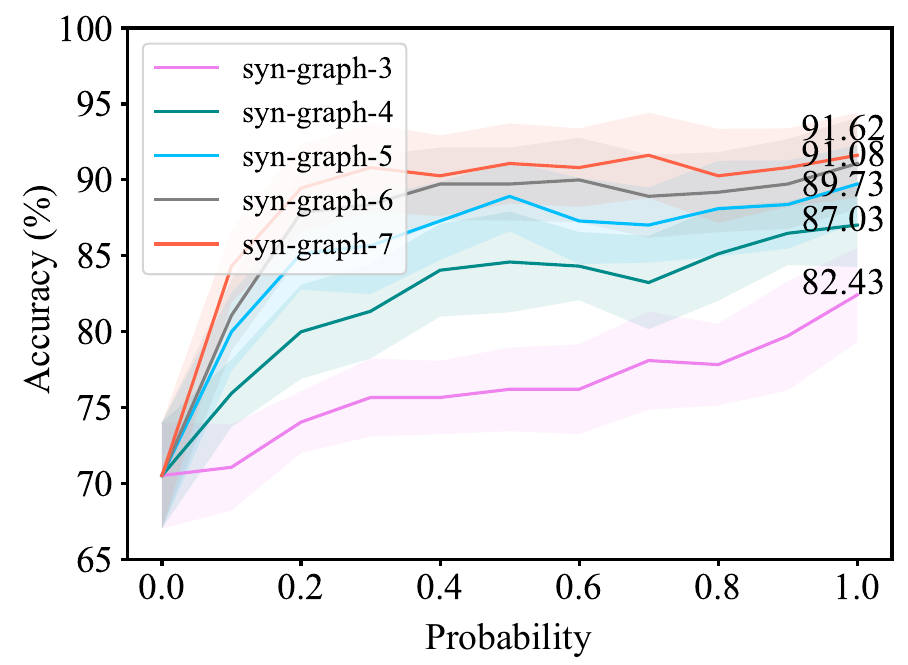}
  }
  \subfigure[]{
    \includegraphics[width=0.3\textwidth]{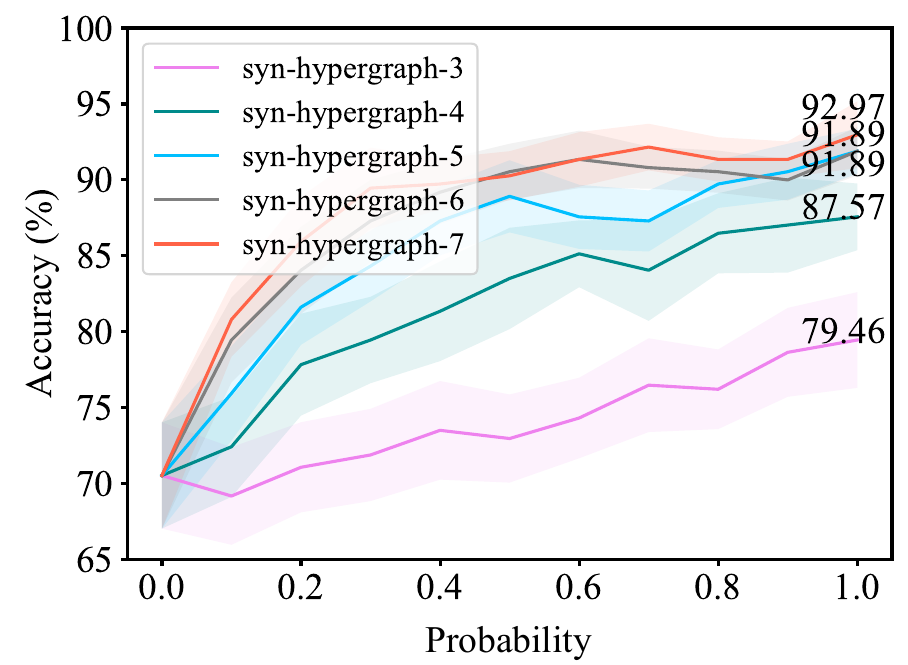}
}
  \caption{\textbf{Experiments on synthetic datasets based on \texttt{Texas}.} The horizontal axis denote the probability of adding node to an existing edge. The numbers attached to the curves denote the corresponding values of probability one. The numerical value indicated in the legend of (a)(c), e.g., ``syn-hypergraph-3'', corresponds to the rank of each hyperedge in the synthetic hypergraphs. The numerical value indicated in the legend of (b), e.g., ``syn-graph-3'', corresponds to the graph obtained from the corresponding synthetic hypergraph, as stated in the text. The homophily score is calculated based on the clique expansion of hypergraphs.}
  \label{fig:q5-5}
\end{figure}

The proposed framework utilizes a universal architecture for both graphs and hypergraphs with difference lying in the construction of the $\mathbf{P}_\mathcal{E}$ matrix. $\mathbf{P}_\mathcal{E}$ contains at most two nonzero elements per row for graphs, whereas for hypergraphs it contains three or more nonzero elements per row. In practise, an important question is the conversion between graphs and hypergraphs. To guarantee as high homophilic extent as possible when synthesizing hypergraphs from graphs, a technique is to add certain node to existing edge with a probability, which has same label with at least one of the original nodes in the edge. Our experiments on synthesized hypergraphs show that this approach significantly increases the homophilic extent, leading to improved performance, see Figure \ref{fig:q5-5}(a). We also provide the corresponding homophily scores for different probabilities of adding node to an existing edge and different ranks of synthetic hypergraphs in Table \ref{table: ab-6}.

\begin{table}[htbp]
  \centering
  \caption{\textbf{Corresponding homophily scores for different probabilities of adding node to an existing edge and different ranks of synthetic hypergraphs.}}
  \label{table: ab-6}
  \resizebox{0.8\textwidth}{!}{%
  \begin{tabular}{cccccccccccc}
    \toprule
    Probability &  0.0 &  0.1 &  0.2 & 0.3  & 0.4  & 0.5  & 0.6  & 0.7  & 0.8  & 0.9  & 1.0\\
    \midrule
    Rank 3    & 0.06 & 0.14 & 0.21 & 0.25 & 0.28 & 0.30 & 0.32 & 0.34 & 0.35 & 0.36 & 0.37 \\
    Rank 4    & 0.06 & 0.26 & 0.36 & 0.42 & 0.44 & 0.46 & 0.49 & 0.50 & 0.51 & 0.52 & 0.53 \\
    Rank 5    & 0.06 & 0.37 & 0.48 & 0.54 & 0.56 & 0.57 & 0.60 & 0.60 & 0.61 & 0.62 & 0.62 \\
    Rank 6    & 0.06 & 0.46 & 0.57 & 0.62 & 0.64 & 0.65 & 0.66 & 0.66 & 0.67 & 0.68 & 0.68 \\
    Rank 7    & 0.06 & 0.54 & 0.64 & 0.68 & 0.69 & 0.70 & 0.71 & 0.71 & 0.72 & 0.72 & 0.73 \\
    \bottomrule
  \end{tabular}
  }
\end{table}

Moreover, to compare the performance of our UniG-Encoder on graphs and hypergraphs that have same homophilic extent, based on the above synthesized hypergraph datasets, we obtain the graph datasets with same homophilic extent by adding new edges that belong to the clique expansion of the corresponding hyperedges in the synthesized hypergraph datasets. Here to ensure a fair comparison, we fix the training hyperparameters, such as the learning rate of 0.001 and the hidden dimension of 64. Results in Figure \ref{fig:q5-5}(b)(c) show that our framework performs better on hypergraphs than graphs with the same high homophily and performs better on graphs than hypergraphs with the same low homophily, which are also reflected in Table \ref{table: ab-7}.

\begin{table}[htbp]
  \centering
  \caption{\textbf{Performance on synthetic graph and hypergraph datasets with different probability of adding node to an existing edge.}}
  \label{table: ab-7}
  \resizebox{0.8\textwidth}{!}{%
  \begin{tabular}{cccccc}
    \toprule
    Probability                &  0.0 &  0.1 &  0.2 & 0.3  & 0.4  \\
    \midrule
    syn-graph-3       & $ 70.54\pm 6.99 $ & $71.08 \pm5.68 $ &$ 74.05\pm4.05 $ & $75.68 \pm5.13 $ & $75.68 \pm4.83 $ \\
    syn-hypergraph-3
    & $70.54 \pm 6.99$ & $69.19 \pm 6.42$ & $ 70.08\pm 5.93 $ & $ 71.89\pm 6.07 $ & $ 73.51\pm 6.49 $\\

    syn-graph-4       & $70.54 \pm6.99 $ & $75.95\pm4.43 $ & $ 80.00\pm6.19 $ & $ 81.35\pm6.22 $ & $84.05 \pm6.10 $ \\
    syn-hypergraph-4
    & $ 70.54\pm6.99 $ & $ 72.43\pm 6.60$ & $ 77.84\pm6.71 $ & $ 79.46\pm5.69 $ & $ 81.35\pm6.56 $ \\
    syn-graph-5             & $70.54 \pm6.99 $ & $ 80.00\pm5.16 $ & $ 85.13\pm4.72 $ & $ 85.67\pm6.40 $ & $ 87.30\pm5.14 $ \\

    syn-hypergraph-5

    & $70.54 \pm6.99 $ & $ 75.95\pm6.22 $ & $81.62 \pm4.95 $ & $84.32 \pm4.65 $ & $87.30 \pm4.53 $ \\

    syn-graph-6
    & $70.54 \pm6.99 $ & $ 81.08\pm4.68 $ & $ 87.84\pm5.57 $ & $ 88.38\pm6.40 $ & $ 89.73\pm4.80 $ \\
    syn-hypergraph-6

    & $70.54 \pm6.99 $ & $ 79.46\pm5.57 $ & $ 84.05\pm5.73 $ & $ 87.30\pm5.41 $ & $ 89.19\pm4.19 $ \\

    syn-graph-7  & $70.54 \pm6.99 $ & $ 84.32\pm4.80 $ & $ 89.46\pm5.73 $ & $90.81 \pm5.82 $ & $90.27 \pm5.30 $ \\
    syn-hypergraph-7

    & $70.54 \pm6.99 $ & $ 80.81\pm4.90 $ & $ 85.95\pm5.90 $ & $ 89.46\pm 5.33$ & $ 89.73\pm3.38 $ \\
    \bottomrule
    \toprule
    0.5  & 0.6  & 0.7  & 0.8  & 0.9  & 1.0 \\
    \midrule
    $76.22 \pm5.51 $ & $ 76.22\pm5.90 $ & $ 78.11\pm6.45 $ & $ 77.84\pm5.38 $ & $ 79.73\pm7.18 $ & $ 82.43\pm6.19 $\\

    $ 72.97\pm 5.80 $ & $ 74.32 \pm 5.30 $ & $ 76.49\pm 6.17 $ & $ 76.22\pm 5.24 $ & $78.65 \pm 5.85 $ & $ 79.46\pm6.30 $ \\

    $84.59 \pm6.63 $ & $84.32 \pm4.49 $ & $ 83.24\pm6.14 $ & $85.13 \pm6.19 $ & $86.49 \pm4.19 $ & $87.03 \pm5.51 $\\

    $83.51 \pm6.67 $ & $85.13 \pm4.40 $ & $ 84.05\pm 6.67$ & $86.49 \pm5.27 $ & $ 87.03\pm6.26 $ & $87.57 \pm4.39 $\\

    $88.92 \pm4.59 $ & $ 87.30\pm5.68 $ & $87.03 \pm4.95 $ & $88.11 \pm6.30 $ & $ 88.38\pm5.80 $ & $ 89.73\pm5.10 $\\

    $88.92 \pm4.75 $ & $87.57 \pm4.22 $ & $87.30 \pm4.02 $ & $89.73 \pm3.15 $ & $90.54 \pm3.68 $ & $91.89 \pm2.69 $\\

    $ 89.73\pm4.80 $ & $ 90.00\pm5.55 $ & $ 88.92\pm 5.47$ & $ 89.20\pm5.27 $ & $ 89.73\pm5.90$ & $ 91.08\pm6.05 $\\

    $ 90.54\pm3.68 $ & $ 91.35\pm3.78 $ & $ 90.81\pm2.76 $ & $ 90.54\pm2.77 $ & $ 90.00\pm2.72 $ & $ 91.89\pm 3.20$\\

    $ 91.08\pm5.28 $ & $ 90.81\pm5.16 $ & $ 90.62\pm5.60 $ & $90.27 \pm6.19 $ & $ 90.81\pm5.16 $ & $91.62 \pm5.60 $\\

    $ 90.27\pm3.24 $ & $ 91.35\pm3.59 $ & $92.16 \pm3.07 $ & $ 91.35\pm 2.91$ & $ 91.35\pm2.36 $ & $ 92.97\pm 4.39$\\
    \bottomrule
  \end{tabular}
  }
\end{table}

\section*{Declaration of competing interest}

The authors declare that they have no known competing financial interests or personal relationships that could have appeared to influence the work reported in this paper.

\section*{Data availability}

Data will be made available on request.

\section*{Acknowledgements}

This work is supported by the National Natural Science Foundation of China (No. 12101133) and Shanghai Sailing Program (No. 21YF1402300). This work is also supported by Shanghai Municipal Science and Technology Major Project (No. 2021SHZDZX0103).

\bibliography{Ref}

\end{document}